\documentclass[11pt,a4paper]{article}
\usepackage[hyperref]{acl2021}
\usepackage{times}
\usepackage{latexsym}
\usepackage{graphicx}
\usepackage{amsmath}

\usepackage{float}
\usepackage{xcolor}
\usepackage[normalem]{ulem} 
\newcommand\green{\bgroup\markoverwith
{\textcolor{green}{\rule[-.5ex]{2pt}{2.5ex}}}\ULon}
\newcommand\red{\bgroup\markoverwith
{\textcolor{red}{\rule[-.5ex]{2pt}{2.5ex}}}\ULon}
\usepackage{microtype}

\aclfinalcopy % Uncomment this line for the final submission
\def\aclpaperid{1586} %  Enter the acl Paper ID here

\title{Entity Concept-enhanced Few-shot Relation Extraction}

\author{

ShanYang\textsuperscript{\rm 1},
Yongfei Zhang\textsuperscript{\rm 1,2,3}\thanks{\ \ Corresponding Author}  ,
Guanglin Niu\textsuperscript{\rm 1},
Qinhua Zhao\textsuperscript{\rm 4},
Shiliang Pu\textsuperscript{\rm 5}
\\ % All authors must be in the same font size and format. Use \Large and \textbf to achieve this result when breaking a line
\textsuperscript{\rm 1}Beijing Key Laboratory of Digital Media, School of Computer Science and Engineering,
\\BeiHang University, Beijing 100191, China\\
\textsuperscript{\rm 2}State Key Laboratory of Virtual Reality Technology and Systems,\\ BeiHang University,Beijing 100191, China \\
\textsuperscript{\rm 3}Pengcheng Laboratory, Shenzhen 518055, China\\
\textsuperscript{\rm 4}State Key Laboratory of Software Development Environment
School of Computer Science\\ and Engineering, BeiHang University, Beijing 100191, China\\
\textsuperscript{\rm 5}Hikvision Research Institute, Hangzhou 311500, China
\\ %If you have multiple authors and multiple affiliations
\{shanyang, yfzhang, beihangngl, zhaoqh\}@buaa.edu.cn, pushiliang.hri@hikvision.com
}

\date{}

\begin{document}
\maketitle
\begin{abstract}
Few-shot relation extraction (FSRE) is of great importance in long-tail distribution problem, especially in special domain with low-resource data. Most existing FSRE algorithms fail to accurately classify the relations merely based on the information of the sentences together with the recognized entity pairs, due to limited samples and lack of knowledge. To address this problem, in this paper, we proposed a novel entity CONCEPT-enhanced FEw-shot Relation Extraction scheme (ConceptFERE), which introduces the inherent concepts of entities to provide clues for relation prediction and boost the relations classification performance. Firstly, a concept-sentence attention module is developed to select the most appropriate concept from multiple concepts of each entity by calculating the semantic similarity between sentences and concepts. Secondly, a self-attention based fusion module is presented to bridge the gap of concept embedding and sentence embedding from different semantic spaces. Extensive experiments on the FSRE benchmark dataset FewRel have demonstrated the effectiveness and the superiority of the proposed ConceptFERE scheme as compared to the state-of-the-art baselines. Code is available
at \textit{\textcolor{red}{https://github.com/LittleGuoKe/ConceptFERE}}.
\end{abstract}
\section{Introduction}
Relation extraction (RE) is a fundamental task for knowledge graph construction and inference, which however often encounters challenges of long-tail distribution and low-resource data, especially in practical applications including medical or public security fields. In this case, it is difficult for existing RE models to learn effective classifiers \citep{zhang2019long,han2020more}. Therefore, FSRE has become a hot topic in both academia and industry. Existing FSRE methods can be roughly divided into two categories according to the type of adopted training data. The models of the first category only uses the plain text data, without any external information. 
% Use metric learning for few-shot relation extraction\citep{koch2015siamese,snell2017prototypical}, their purpose is to learn general representation and metric function.
The representative Siamese \citep{koch2015siamese} and Prototypical \citep{snell2017prototypical} network in metric learning are used in the FSRE task to learn representation and metric function. BERT-PAIR \citep{gao2019fewrel} pairs up all supporting instances with each query instance, and predicts whether the pairs are of the same category, which can be regarded as a variant of the Prototypical network. Gao \citep{gao2019hybrid} and Ye \citep{ye2019multi} add the attention mechanism to enhance the prototype network. In order to alleviate the problem of insufficient training data, MICK \citep{geng2020mick} learns general language rules and grammatical knowledge from cross-domain datasets. Wang \citep{wang2020learning} proposes the CTEG model to solve the relation confusion problem of FSRE. Cong \citep{cong2020inductive} proposes an inductive clustering based framework, DaFeC, to solve the problem of domain adaptation in FSRE.
\begin{table*}
\centering
\begin{tabular}{lll}
\hline 
Relation  & founder\\
Sentence & \textbf{Microsoft} was founded by \textbf{Bill Gates} and Paul Allen on April 4, 1975 \\
Head entity concept & company, vendor, client\\
Tail entity concept & person, billionaire, entrepreneur\\
\hline
\end{tabular}
\caption{The bold words in the sentence correspond to the head entity and the tail entity.}
\end{table*}

Since the information of the plain text is limited in FSRE scenarios, the performance gain is marginal. Thus, the algorithms in the second category introduce external information, to compensate the limited information in FSRE, so as to enhance the performance. In order to improve the model's generalization ability for new relations, Qu \citep{qu2020few} studies the relationship between different relations by establishing a global relation graph. The relations in the global relation graph come from Wikidata.\footnote{https://www.wikidata.org/} TD-Proto \citep{yang2020enhance} introduces text descriptions of entities and relations from Wikidata to enhance the prototype network and provide clues for relation prediction.

% When doing relation extraction, the relation is classified according to the information of the sentence and the entity pair. \citep{han2020more} tested on the common relation extraction datasets only uses the entity name and not uses the entire sentence for relation extraction, which can achieve good results. This shows that the entity name is very important for relation extraction. However, most of the existing work only uses the name of the entity to provide information about the entity, which is not enough. Given the entity definition, the relation corresponding to the entity pair in the sentence is limited to a range, which can reduce the relation category that the model needs to predict and improve the classification performance. The definition of an entity can generally come from the concept in the knowledge graph or the text description in the Wikipedia. 
Although the introduction of knowledge of text description can provide external information for FSRE and achieve state-of-the-art performance, TD-proto only introduces one text description in Wikidata for each entity. However, this might suffer from the mismatching between entity and text description and leads to the degraded performance. Besides, since the text description for each entity is often relatively long, it is not a easy job to extract the most useful information within the long text description. 

In contrast to the long text descriptions, the concept is an intuitive and concise description of an entity and can be readily obtained from concept databases, like YAGO3, ConceptNet and Concept Graph, etc. Besides, the concept is more abstract than the specific text description for each entity, which is an idea compensation to the limited information in FSRE scenarios.

% However, various concepts of entities can be obtained from Goncept Graph, and we can select concepts suitable for the current sentence from multiple concepts to solve the above problems. Since concepts are more concise and abstract than text descriptions, different entities may correspond to different descriptions but can correspond to the same concepts, resulting in the generalization ability of concepts is stronger.

As shown in Table 1, intuitively knowing that the concept of head entity is a \textit{company} and the concept of tail entity is an \textit{entrepreneur}, the relation corresponding to the entity pair in the sentence can be limited to a range: \textit{ceo}, \textit{founder}, \textit{inauguration}. On the other hand, some relations should be wiped out, e.g., \textit{educated at}, \textit{presynaptic connection}, \textit{statement describes}. The semantic information of concept can assist determining the relation: \textit{founder} predicted by the model.

To address the above challenges, we propose a novel entity CONCEPT-enhanced FEw-shot Relation Extraction scheme (ConceptFERE), which introduces the entity concept to provide effective clues for relation prediction. \textit{Firstly}, as shown in Table 1, one entity might have more than one concept from different aspects or hierarchical levels and only one of the concepts might be valuable for final relation classification. Therefore, we design a concept-sentence attention module to choose the most suitable concept for each entity by comparing the semantic similarity of the sentence and each concept. \textit{Secondly}, since the sentence embedding and pre-trained concept embedding are not learned in the same semantic space, we adopt the self-attention mechanism \citep{devlin2018bert} for word-level semantic fusion of the sentence and the selected concept for final relation classification. Experimental results on benchmark dataset show that our method achieves state-of-the-art FSRE performance.

\section{Model}
\subsection{System Overview}
Figure 1 shows the structure of our proposed ConceptFERE. The sentence representation module uses BERT to obtain the sentence embedding, the concept representation adopts the pre-trained concept embedding \citep{shalaby2019beyond}, which uses the skip-gram model to learn the representation of the concept on the Wikipedia text and the Concept Graph. Relation classifier can be implemented by the fully connected layer. The remaining modules of the model will be described in detail below.

\begin{figure*}
    \centering
    \includegraphics[scale=0.248]{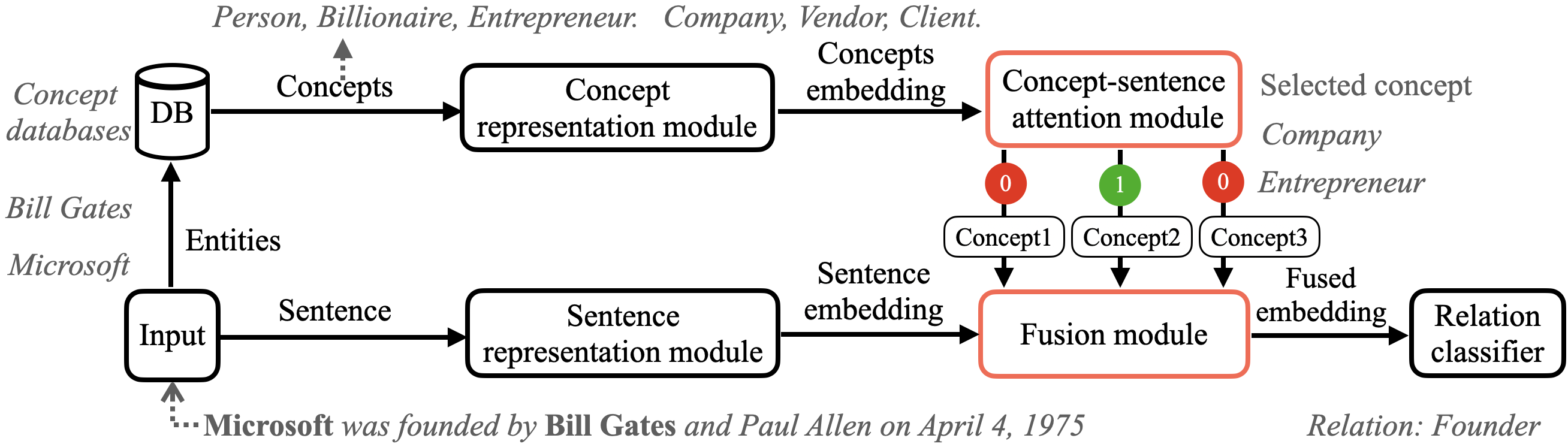}
    \caption{Structure diagram of ConceptFERE model.}
\end{figure*}

% \subsection{ConceptFERE}
% When ConceptFERE(Simple) selects the entity concept, concepts that have nothing to do with sentence semantics may also be given a lower similarity value, and they will still participate in relation extraction, which is unreasonable. To select the appropriate concept, we design the concept selection module. Then use the self-attention based fusion module to merge the concept and sentence semantically at the word-level, as shown in Figure 1.
\subsection{Concept-Sentence Attention Module}
Intuitively, one needs to pay more attention to the concept of high semantic correlation with the sentence, which can provide more effective clues for RE. Firstly, since the pre-trained concept embedding ($v_{c}$) and sentence embedding ($v_{s}$) are not learned in the same semantic space, we can not compare the semantic similarity directly. So the semantic transformation is performed by multiplying the $v_{c}$ and $v_{s}$ by the projection matrix $P$ to get their representations $v_{c}P$ and $v_{s}P$ in the same semantic space, where $P$ can be learned by fully connected networks. Secondly, by calculating the semantic similarity between sentence and each concept of entity, the similarity value is obtained 
% as follow:
% \begin{equation}
% \operatorname{sim_{i}}=\left\langle V_{i}, V_{s}\right\rangle
% \end{equation}
% where
the dot product of the concept embedding $v_{c}$ and the sentence embedding $v_{s}$ as similarity $sim_{cs}$.
Finally, in order to select a suitable concept from the calculated similarity value, we design the 01-GATE. The similarity value is normalized by the Softmax function.
% in the 01-GATE as follow:
% \begin{equation}
% \operatorname{softmax}\left(sim_{i}\right)=\frac{e^{\operatorname{sim}_{i}}}{\sum_{k=1}^{K} e^{sim _{k}}}
% \end{equation}
% where\begin{word}
% K
% \end{word}represents the number of concepts corresponding to each entity. 
If $sim_{cs}$ is less than the set threshold $\alpha$, 01-GATE assigns 0 to the attention score of the corresponding concept, and this concept will be excluded in subsequent relation classification. We choose the suitable concept with the attention score of 1, which is used as a effective clue to participate in relation prediction.

\subsection{Self-Attention based Fusion Module}
% As discussed in section 1, given the entity concept, the relation corresponding to the entity pair will be limited to a range. In order to determine the relation category that the model needs to predict, we can combine the semantic information of sentences with the concept of entities. Meanwhile, 
Since concept embedding and the embedding of words in sentences are not learned in the same semantic space, we design a self-attention \citep{devlin2018bert} based fusion module to perform word-level semantic fusion of the concept and each word in the sentence. First, the embedding of all words in the sentence and the selected concept embedding are concatenated, 
% as follow:\begin{equation}
% S C=\left[W_{1}, W_{2}, \cdots, W_{m}, C_{1}, \cdots, C_{k}\right]
% \end{equation}
and then fed to the self-attention module. As shown in Figure 2, the self-attention module calculates the similarity value between the concept and each word in the sentence. It multiplies the concept embedding and the similarity value, and then combine with its corresponding word embedding as follow:
% \begin{equation}
% \text{$fusion$}_{v_{i}}=\sum_{j=1}^{N} \operatorname{sim}\left(q_{i}, k_{j}\right) v_{j}
% \end{equation}
\begin{equation}
\text{$fusion$}_{v_{i}}=\sum_{j=1}^{N} sim\left(q_{i}, k_{j}\right) v_{j}
\end{equation}
where $fusion_{v_{i}}$ represents the embedding of $v_{i}$ after $v_{i}$ performs the word-level semantic fusion. The $q_{i}$, $k_{j}$, and $v_{j}$ are derived from self-attention, they represent the concept embedding or the word embedding.
% \subsection{ConceptFERE(Simple)}
% In this section, we propose a simplified version of the model, without carefully designed conceptselection and fusion module, we call ConceptFERE(Simple).
% % , which is very convenient to apply to other natural language processing tasks.
% We only need to concatenate the concepts and sentences and input them into the existing model. Specifically, we can input the concatenated sentences and concepts into BERT-PAIR \citep{gao2019fewrel}. 
% The self-attention in BERT-PAIR selects the appropriate concept by calculating the semantic similarity between the concept and the sentence, the fusion process of concepts and sentences is the same as that described in section 2.3.

\begin{figure}
    \centering
    \includegraphics[scale=0.248]{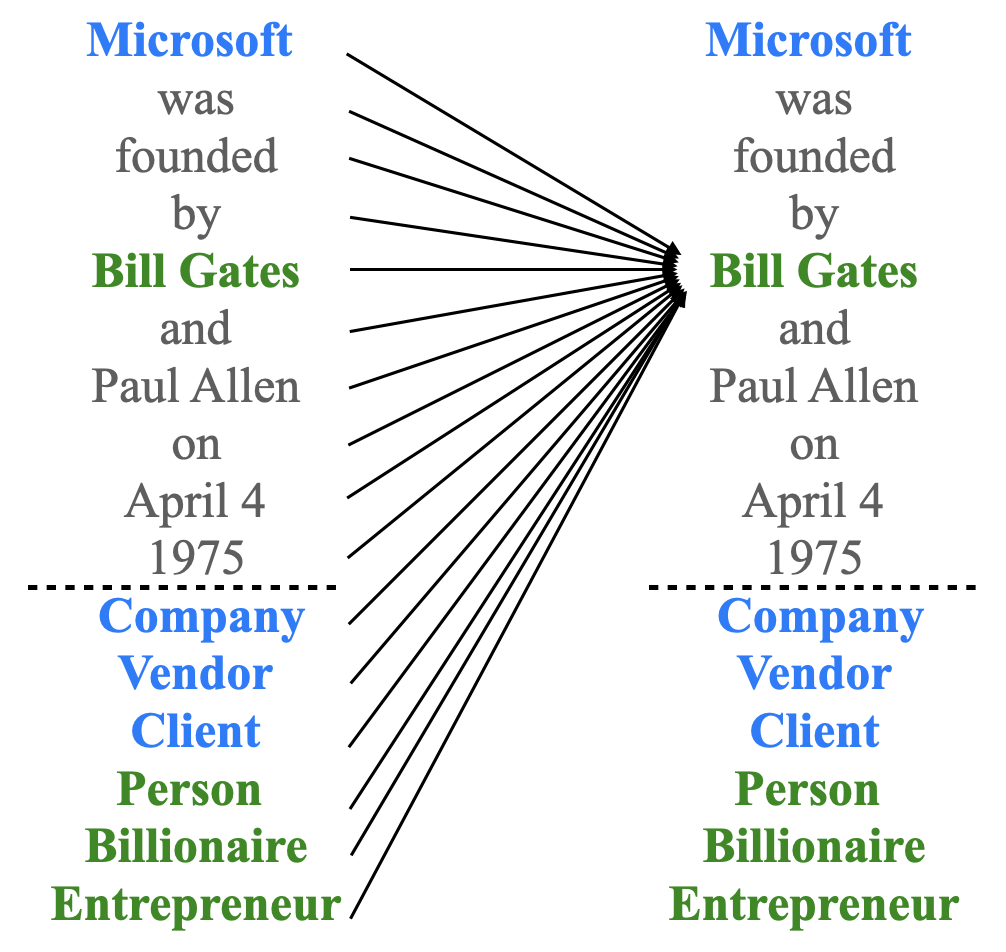}
    \caption{The word-level semantic fusion.}

\end{figure}

\section{Experiment}
% highlight our model advantagde in few shot
\subsection{Dataset, Evaluation and Comparable Models}
\textbf{Dataset}: In order to verify our proposed method, we use the most commonly used FSRE dataset FewRel \citep{han2018fewrel}, which contains 100 relations and 70,000 instances extracted from Wikipedia, with 20 relations in the unpublished test set. So we follow previous work \citep{yang2020enhance} to re-split the published 80 relations into 50, 14 and 16 for training, validation and testing, respectively.

% Evaluation: in N-way-K-shot scenario, accuracy is used as the performance metric.
\textbf{Evaluation}: N-way-K-shot (N-w-K-s) is commonly used to simulate the distribution of FewRel in different situations, where N and K denote the number of classes and samples from each class, respectively. In N-w-K-s scenario, accuracy is used as the performance metric.

\textbf{Comparable Models}: We choose excellent baseline models, GNN \citep{garcia2017few}, SNAIL \citep{mishra2017simple}, Proto \citep{snell2017prototypical}, HATT-Proto \citep{gao2019hybrid}, MLMAN \citep{ye2019multi} and TD-proto \citep{yang2020enhance} for comparison, and their experimental results are derived from \citep{yang2020enhance}.
\subsection{Model Training Details}
 The BERT parameters are initialized by bert-base-uncased, and the hidden size is 768. The threshold $\alpha$ is 0.7. Hyperparameters such as learning rate follow the settings in \citep{gao2019fewrel}. The entity concept is obtained from Concept
\begin{table*}
\centering
\begin{tabular}{llllll}
\hline 
\textbf{Model} &\textbf{Encoder}&\textbf{5-w-1-s}&\textbf{5-w-5-s}&\textbf{10-w-1-s}&\textbf{10-w-5-s}\\
GNN \citep{garcia2017few} &  CNN  & 67.30   & 78.84   & 54.10   & 62.89 \\
SNAIL \citep{mishra2017simple}  &   CNN  &  71.13  &   80.04  &   50.61  &   66.68 \\
Proto \citep{snell2017prototypical}   &    CNN   &   74.29   &    85.18  &    61.15  &    74.41 \\
HATT-Proto \citep{gao2019hybrid}   &    CNN   &   74.84   &    85.81 & 62.05 & 75.25 \\
MLMAN \citep{ye2019multi} & CNN & 78.21 & 88.01 & 65.70 & 78.35 \\
Bert-PAIR \citep{gao2019fewrel} & BERT & 82.57 & 88.47 & 73.37 & 81.10 \\
\hline
TD-Proto \citep{yang2020enhance} & BERT & 84.76 & 92.38 & 74.32 & 85.92 \\
\hline
ConceptFERE & BERT & \textbf{89.21} & -- & \textbf{75.72} & -- \\
ConceptFERE (Simple) & BERT & 84.28 & 90.34 & 74.00 & 81.82 \\ 
\hline
\end{tabular}
\caption{Accuracies (\%) of different models on test set.}
\end{table*}
 Graph\footnote{https://concept.research.microsoft.com/Home/Download}. Concept
 Graph is a large-scale common sense conceptual knowledge graph developed by Microsoft, which contains concept of entities stored in triplets (Entity, IsA, Concept) and can provide concept knowledge for entities in ConceptFERE. The concept embedding adopts the pre-trained concept embedding\footnote{https://sites.google.com/site/conceptembeddings/} \citep{shalaby2019beyond}.

Our proposed scheme is implemented on top of BERT-PAIR, since the concept provided by ConceptFERE can be used as an effective clue.

% \begin{table*}
% \centering
% \begin{tabular}{p{100pt}p{340pt}}
% \hline 
% Sentence & In 2015 , Garland made his directorial debut with \textbf{\textit{Ex Machina}}, a science fiction thriller which explores the relationship between mankind and \textbf{artificial intelligence}.\\
% Head entity concept &
% \red {technology}, \red{field},\green{topic}\\
% Tail entity concept & \green{film}, \green{science fiction movie}, \red{project}\\
% Relation & main subject\\
% \hline
% \end{tabular}
% \caption{\label{font-table}Visualization attention score of the entity concept. The red and green backgrounds indicate that the attention score is 0 and 1, respectively. For each entity, we select three concepts from Concept Graph. Bold fonts are head entity and tail entity, respectively}
% \end{table*}

% \begin{table*}
% \centering
% \begin{tabular}{p{100pt}p{340pt}}
% \hline 
% Sentence & In 1937 , \textbf{\textit{Savitri}} was produced in \textbf{Hindi} directed by Franz Osten . \\
% Head entity concept &
% \green{serial}, \red{high yielding variety}, \red{chief god}\\
% Tail entity concept &\green{language}, \green{indo aryan language}, \green{indian language}\\
% Relation & original language of film or TV show\\
% \hline
% \end{tabular}
% \caption{\label{font-table}Visualization attention score of the entity concept. The red and green backgrounds indicate that the attention score is 0 and 1, respectively. For each entity, we select three concepts from Concept Graph. Bold fonts are head entity and tail entity, respectively}
% \end{table*}

\subsection{Performance and Comparisons}
Table 2 tabulates the performance of different comparable models on the test set, where the algorithms in the first group are those state-of-the-art schemes without using any external information, while the TD-Proto in the second group uses external information of text descriptions of entities, and finally our proposed scheme in the third group. It should be noted that, due to the insufficient computing power of our GPU, the performance of the proposed ConceptFERE scheme is tested only under 5way1shot and 10way1shot scenarios. It can be observed from Table 2 that the proposed ConceptFERE model achieves the best performance, as compared to all the comparable schemes. More specifically,  ConceptFERE achieves respectively 4.45 and 1.4 gains over the latest TD-Proto using external entity descriptions. And a performance gain of 6.64 and 2.35 is registered as compared to Bert-PAIR, the best model in the first category, under the 5way1shot and 10way1shot scenarios, respectively. This might due to that the generalization ability of concepts is stronger than text description and it is more suitable for FSRE. In theory, 1-shot relation extraction is a more difficult task than 5-shot relation extraction. The experimental results of 1-shot relation extraction have illustrated the effectiveness and superiority of our approach. We believe that our ConceptFERE scheme would also achieve the best performance under the other two scenarios.
%  It can be observed from Table 2 that the performance of the model can be greatly improved after the BERT is introduced. So in the FSRE scenario, BERT pre-trained on the large-scale corpus can provide better representation and prior knowledge. The experimental results of the BERT-PAIR ConceptFERE(Simple) model are better than BERT-PAIR, which proves that concepts can provide effective clues for relation prediction. Due to the insufficient computing power of the GPU, we did not do part of the experiment. Marked with the ‘--’ symbol in the table. Our method slightly surpasses the SOTA model TD-proto. It may be that we have solved the problem of mismatch, and the generalization ability of concepts is stronger than text description and it is more suitable for FSRE.

% \begin{table}
% \centering
% \begin{tabular}{llllll}
% \hline 
% \textbf{Model} &\textbf{5 way 1 shot}}\\
% \hline
% \textbf{ConceptFERE(bert)} &\textbf{89.21}}\\
% \hline
% \textbf{w/o FUSION}  &\textbf{83.11}}\\
% \textbf{w/o ATT}  &\textbf{84.03}}\\
% \textbf{w/o ATT $\&$ FUSION}  &\textbf{82.57}}\\
% \hline
% \end{tabular}
% \caption{\label{font-table}Results of ablation study with ConceptFERE.}
% \end{table}

\begin{table}
\centering
\begin{tabular}{llllll}
\hline 
\textbf{Model} &\textbf{5-w-1-s}\\
\hline
ConceptFERE (bert) &89.21\\
\hline
w/o FUSION  & 83.11\\
w/o ATT & 84.03\\
w/o ATT $\&$ FUSION  &82.57\\
\hline
\end{tabular}
\caption{Results of ablation study with ConceptFERE.}
\end{table}

\subsection{Ablation Study}
% In this section, We explore the effectiveness of the modules proposed above. As shown in Table 4, The B-P SC model simply uses the concept without using  concept-sentence attention and fusion module, and the model performance drops sharply. This can reflect that the fusion module and module can effectively select appropriate concepts and perform word-level semantic integration of concepts and sentences.
In this section, to verify the effectiveness of the proposed concept-sentence attention module and self-attention based fusion module, presented in 2.2 and 2.3, respectively.
% The self-attention in BERT-PAIR selects the appropriate concept by calculating the semantic similarity between the concept and the sentence, the fusion process of concepts and sentences is the same as that described in Section 2.3.  
As shown in Table 3, without using the concept-sentence attention and fusion module, the model performance of ConceptFERE (simple) drops sharply. This proves that the proposed concept-sentence attention module (ATT) and fusion module (FUSION) can effectively select appropriate concepts and perform word-level semantic integration of concepts and sentences. On the other hand,  we present a simplified version of the ConceptFERE model, denoted as ConceptFERE (Simple), in which both the concept selection and fusion module are removed and the concepts and sentences are concatenated and inputted into the relation classification model. Specifically, we can input the concatenated sentences and concepts into BERT-PAIR \citep{gao2019fewrel}. As shown in Table 2, ConceptFERE (simple) achieves much better performance as compared to Bert-PAIR, the best model in the first category, under all four scenarios. This further validates the effectiveness of introducing the concept in enhancing the RE performance. More importantly, it can be easily applied to other models. As mentioned above, we only need to input the concatenated entity concepts and sentences into the model.

\section{Conclusion}
In this paper, we have studied the FSRE task and presented a novel entity concept-enhanced FSRE scheme (ConceptFERE). The concept-sentence attention module was designed to select the appropriate concept from multiple concepts corresponding to each entity, and the fusion module was designed to integrate the concept and sentence semantically at the word-level. The experimental results have demonstrated the effectiveness of our method against state-of-the-art algorithms. As a future work, the commonsense knowledge of the concepts as well as the possible relations between them will be explicitly considered to further enhance the FSRE performance.

\section*{Acknowledgments}
This work was partially supported by the National
Natural Science Foundation of China (No.
61772054, 62072022), and the NSFC Key Project
(No. 61632001) and the Fundamental Research
Funds for the Central Universities.

\bibliographystyle{acl_natbib}
\bibliography{acl2021}

%\appendix

\end{document}